# Optimization-based Alignment for Strapdown Inertial Navigation System: Comparison and Extension

Lubin Chang, *Member, IEEE*, Jingshu Li, and Kailong Li

*Abstract*—In this paper, the optimization-based alignment (OBA) methods are investigated with main focus on the vector observations construction procedures for the strapdown inertial navigation system (SINS). The contributions of this study are twofold. First the OBA method is extended to be able to estimate the gyroscopes biases coupled with the attitude based on the construction process of the existing OBA methods. This extension transforms the initial alignment into an attitude estimation problem which can be solved using the nonlinear filtering algorithms. The second contribution is the comprehensive evaluation of the OBA methods and their extensions with different vector observations construction procedures in terms of convergent speed and steady-state estimate using field test data collected from different grades of SINS. This study is expected to facilitate the selection of appropriate OBA methods for different grade SINS.

*Index Terms*—Attitude estimation, inertial navigation, initial alignment, velocity integration formula

I. INTRODUCTION

As a dead-reckoning navigation method, the performance of the strapdown inertial navigation system (SINS)

Manuscript received December 19, 2013; revised April 25, 2014, August 23, 2014; released for publication May 10, 2016.
 DOI. No. 10.1109/TAES.2016.130824.
Refereeing of this contribution was handled by A. Dempster. This work was supported, in part, by the National Natural Science Foundation of China (61304241, 61374206, and 41404002).
Authors' addresses: Department of Navigation Engineering, Naval University of Engineering, China, Jiefang Road 717, Qiaokou District,Wuhan, 430033 People's Republic of China. Corresponding author is L.Chang, E-mail: (changlubin@163.com).

relies largely on the accuracy of the initial conditions which are determined by the so called initial alignment [1, 2]. Up to this day, the initial alignment has been one of the most researched topics for SINS and various methods have been proposed to address this problem [3]. Typically, the initial alignment can be achieved through two consecutive stages: coarse alignment and fine alignment. Regarding the fine alignment, consensus has been reached that the Kalman filtering based optimal estimation scheme is the standard method. The Kalman filtering based fine alignment is founded on the linearized system error models and necessitates the coarse alignment to provide roughly known initial attitude to guarantee the validity of such linearization [4, 5]. Unfortunately, the traditional analytic coarse alignment methods are no longer applicable when the carrier is moving or maneuvering.

Recently, an attitude matrix decomposition based coarse alignment method has been proposed for the carrier under swaying or maneuvering conditions [6-11]. By introducing some "fixed" reference frames known as inertial frames, the attitude matrix can be decomposed into three parts: attitude matrix as a function of the carrier's angular rate, attitude matrix as a function of the Earth and transport rates and a constant attitude matrix encoding the transformation between the body and navigation frames at the very start of the initial alignment. The first two attitude matrices can be obtained through the attitude update procedure according to the corresponding angular rates. The constant attitude matrix can be derived using solutions to Wahba's problem based on the constructed vector observations. To this respect, the attitude alignment problem has been transformed into a continuous attitude determination problem [6]. In [6, 8], such initial alignment method is termed as "optimization-based alignment (OBA)" because the Wahba's problem is virtually a constrained least square problem and all the solutions are devoted to minimizing the corresponding cost function in a optimal manner. Since many fruitful algorithms can be readily used to address the attitude determination problem, say Davenport's q method used in [6, 8, 10], the key of determining the constant attitude matrix has been the construction of the vector observations. Since the constant attitude matrix encoding the transformation between the "fixed" body and navigation frames at the very start of the initial alignment, the vector observations expressed in the corresponding reference frames are constructed based on the specific force equation. The specific force equation encodes the relationship between different kinds of acceleration information and therefore, acceleration information based vector observations are firstly constructed and used in the OBA method in [6]. However, there is much external disturbance inherent in the acceleration information, which can degrade the performance of the OBA method significantly. Although the low-pass filter designed in [6] can restrain the disturbance to a certain extent, the parameters design for the low-pass filter based on different maneuvering conditions is a troublesome process. Meanwhile, the disturbance of the accelerometer outputs has not been taken into account in [6]. A straightforward and effective method to restrain disturbance is to integrate the acceleration information over certain time interval, which is just the case in [8, 10].Regarding the length of the integral time interval, there are mainly two integration procedures: one is the integration procedure over the whole alignment time interval in [8] and the other is the integration procedure over certain time interval with fixed length in [10]. It is well known that restraining effect on the disturbance becomes better and better when the integral time interval increases, which seemingly favors the integration procedures in [8]. Moreover, making use of more acceleration information can accelerate the convergent speed of the OBA method. However, when the integral time interval increases, the inertial sensors biases will also cause accumulative errors in the constructed vector observations and in turn, degrade the performance the OBA method. To this respect, the integral time interval should not be too long, which favors the integration procedures in [10] on the other way.

The aforementioned conflicting conclusions demonstrate that many issues need to be further fully addressed for the OBA methods. Motivated by the aforementioned discussion, this paper is devoted to evaluating the performances of the OBA method with different vector observations construction procedures using field test data collected from different grades of SINS. Since the inertial sensors biases, especially the gyroscopes biases, have a much negative effect on the precision of the calculated attitude, the existing OBA methods will be no longer applicable for the low-grade SINS, say micro-electromechanical system (MEMS) technology based low-cost SINS [12, 13], due to their inability to estimate anything other than the attitude. With this consideration, this paper extends the existing OBA methods and investigates an attitude estimation based alignment method which can estimate the gyroscopes biases coupled with the attitude. Since the investigated attitude estimation based alignment method is used to estimate the attitude at current time that is not constant for the in-motion alignment, we term the investigated alignment method as "dynamic OBA" method. In contrast, the traditional OBA methods are used to determine the constant attitude matrix at the very start of the initial alignment and can be viewed as "static OBA" method. The aforementioned different integration procedures have also been investigated and evaluated in the construction of the vector observations for the developed dynamic OBA method.

The rest of the paper is organized as follows. Section II mathematically makes the static OBA method statement with main focus on the vector observation construction methods with different integration procedures. In Section III, the dynamic OBA method able to estimate the gyroscopes biases coupled with the attitude is developed. Meanwhile, the explicit filtering procedure is also presented to solve the established attitude estimation problem. Section IV evaluates the aforementioned OBA methods comprehensively in terms of convergent speed and steady-state estimate using field test data collected from different grades of SINS. Finally, conclusions are drawn in Section V.

## II. STATIC OBA METHOD

### A. Formulation of the Static OBA Method

For the OBA method, the attitude kinematics equation is used directly and there is no need to derive the corresponding angle error models. Therefore, the OBA method can be seen as an analytical method. The ingeniousness of the OBA method mainly manifests in the reapplication of the navigation (attitude, velocity and position) rate equations in a new manner. For the traditional application, the attitude kinematics equation in terms of attitude matrix is given by

$$\dot{C}_b^n = C_b^n \omega_{nb}^b \times \qquad (1)$$

where

$$\omega_{nb}^b = \omega_{ib}^b - C_b^{nT} \omega_{in}^n = \omega_{ib}^b - C_b^{nT} \left( \omega_{ie}^n + \omega_{en}^n \right) \qquad (2)$$

The definitions of the involved reference frames in this paper are the same with those in [8]. $C_b^n$ denotes the attitude matrix from the body frame to the navigation frame, determining the values of which prior to the navigational computation is just the purpose of the initial alignment. $\omega_{ib}^b$ denotes the body angular rate measured by gyroscopes in the body frame, $\omega_{ie}^n$ the earth rotation rate expressed in the navigation frame and $\omega_{en}^n$ the angular rate caused by the linear motion of the carrier.

Denote by $n(0)$ the inertially non-rotating frame that is aligned with the navigation frame at $t_0$, by $b(0)$ the inertially non-rotating frame that is aligned with the body frame at $t_0$, by $n/b(t)$ the navigation/body frame at $t$. According to the chain rule of the attitude matrix, the attitude matrix $C_b^n(t)$ can be decomposed as

$$C_b^n(t) = C_{n(0)}^{n(t)} C_b^n(0) C_{b(t)}^{b(0)} \qquad (3)$$

Since the attitude matrices $C_{n(t)}^{n(0)}$ and $C_{b(t)}^{b(0)}$ can be determined through the attitude update procedure based on the angular rates $\omega_{in}^n$ and $\omega_{ib}^b$, respectively as

$$\begin{aligned} \dot{C}_{b(t)}^{b(0)} &= C_{b(t)}^{b(0)} \omega_{ib}^b \times \\ \dot{C}_{n(t)}^{n(0)} &= C_{n(t)}^{n(0)} \omega_{in}^n \times \end{aligned} \qquad (4)$$

the heart of determining $C_b^n(t)$ is transformed into determining the constant matrix $C_b^n(0)$. According to [6, 8, 10], the calculation of $C_b^n(0)$ is virtually a continuous attitude determination problem using infinite vector observations which are constructed based on the specific force equation. The specific force equation expressed in the navigation frame is given by

$$\dot{v}^n = C_b^n f^b - \left( 2\omega_{ie}^n + \omega_{en}^n \right) \times v^n + g^n \qquad (5)$$

where $v^n$ denotes the ground velocity, $f^b$ the specific force measured by accelerometers in the body frame and $g^n$ the gravity vector.

Substituting (3) into (5) gives

$$\dot{v}^n = C_{n(0)}^{n(t)} C_b^n(0) C_{b(t)}^{b(0)} f^b - \left(2\omega_{ie}^n + \omega_{en}^n\right) \times v^n + g^n \tag{6}$$

Multiplying $C_{n(t)}^{n(0)}$ on both sides and recognizing the resulting equation yield

$$C_b^n(0) C_{b(t)}^{b(0)} f^b = C_{n(t)}^{n(0)} \left(\dot{v}^n + \left(2\omega_{ie}^n + \omega_{en}^n\right) \times v^n - g^n\right) \tag{7}$$

which can also be rewritten in a compact form as

$$C_b^n(0) \alpha_a = \beta_a \tag{8}$$

with

$$\alpha_a = C_{b(t)}^{b(0)} f^b \tag{9a}$$

$$\beta_a = C_{n(t)}^{n(0)} \left(\dot{v}^n + \left(2\omega_{ie}^n + \omega_{en}^n\right) \times v^n - g^n\right) \tag{9b}$$

The determination of the constant matrix $C_b^n(0)$ in (8) based on the vector observations in (9) is well known as the Wahba's problem [14]. Accordingly, many fruitful algorithms have been developed for such problem, such as the three axis attitude determination (TRIAD) based method, Davenport's q method, the quaternion estimator (QUEST) based method and so on. More details about these attitude determination methods can be found in the excellent survey paper [14] and the references therein. In [9], the TRIAD based method is applied to determine the constant matrix $C_b^n(0)$ while in [6, 8, 10, 11], the Davenport's q method is applied. Since many effective attitude determination methods are readily to be used, the construction methods for the vector observations are crucial to such problem. Although the vector observations in (9) can be used directly, the severe disturbances inherent in the acceleration information, say $f^b$ and $\dot{v}^n$, have a much negative effect on the performance of the attitude determination. A straightforward way to attenuate the negative effect of external disturbances is to integrate the acceleration information through certain time interval. According to the different integration procedures, there are mainly two different vector observations constructions methods for the developed attitude determination problem, which will be presented detailedly in the following section.

*B. Velocity Integration Formula for Vector Observations Construction*

Denote by $t$ the current time instant during the initial alignment process. In [8], the vector observations in terms of acceleration in (9) are integrated over the time interval $[0,t]$. The resulting form of the attitude determination problem associated with the vector observations in terms of velocity is given by

$$C_b^n(0)\alpha_v = \beta_v \tag{10}$$

with

$$\alpha_v = \int_0^t C_{b(\tau)}^{b(0)} f^b d\tau \tag{11a}$$

$$\beta_v = \int_0^t C_{n(\tau)}^{n(0)} \left( \dot{v}^n + \left(2\omega_{ie}^n + \omega_{en}^n\right) \times v^n - g^n \right) d\tau \tag{11b}$$

Similarly to [8, 15], denote the current time instant in its discrete-time form as $t \triangleq M\Delta t$ with $\Delta t$ denoting the time duration of the update interval $[t_k, t_{k+1}]$, $k = 0,1,2,\cdots,M-1$. Accordingly, the integration of (11) can be calculated using the velocity integration formula developed in [8, 15] as

$$\alpha_v = \sum_{k=0}^{M-1} C_{b(t_k)}^{b(0)} \int_{t_k}^{t_{k+1}} C_{b(t)}^{b(t_k)} f^b(t) dt \tag{12a}$$

$$\beta_v = C_{n(\tau)}^{n(0)} v^n \Big|_{t_0}^{t} + \sum_{k=0}^{M-1} C_{n(t_k)}^{n(0)} \int_{t_k}^{t_{k+1}} C_{n(t)}^{n(t_k)} \omega_{ie}^n \times v^n dt \\ - \sum_{k=0}^{M-1} C_{n(t_k)}^{n(0)} \int_{t_k}^{t_{k+1}} C_{n(t)}^{n(t_k)} g^n(t) dt \tag{12b}$$

The illustration of such integrating procedure is shown in Fig. 1. It can be seen from Fig. 1 that, as the time goes on, the length of the integration time interval is increasing and, in turn, the attenuating effect of the external disturbances will be more remarkable. Moreover, making use of such integration procedure can guarantee the numbers of the constructed vector observations without loss of intermediate GPS data samples. Since the acceleration information of the accelerometer and the GPS at certain time instant can be used repeatedly for many times, the convergent speed of the attitude determination is expected to be much faster. The downside of such integration procedure is that, the sensor biases will cause accumulative errors in the constructed vector observations as is shown in (12a). In [8, 10], the OBA method is regarded as a coarse alignment method which needs much shorter time interval, so the accumulative errors have an ignored effect on the navigational or more precise SINS. However, for the low-cost SINS, say MEMS based SINS, the negative effect of these accumulative errors will be much remarkable and even invalidate the OBA method.

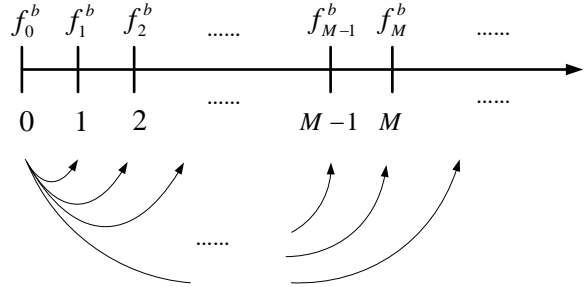

Fig. 1. Integration procedure in (11)

A straightforward way to attenuate the negative effect of these accumulative errors is to integrate the acceleration based vector observations in (9) over time interval $[t_m, t]$ with $t_m \in [0, t]$. The so called interleaved integrating method proposed in [10] is just with such category. In this case, the corresponding attitude determination formulation with the associated vector observations in terms of velocity increment is given by

$$C_b^n(0)\alpha_{\Delta v} = \beta_{\Delta v} \quad (13)$$

with

$$\alpha_{\Delta v} = \int_{t_m}^{t} C_{b(\tau)}^{b(0)} f^b d\tau \quad (14a)$$

$$\beta_{\Delta v} = \int_{t_m}^{t} C_{n(\tau)}^{n(0)} \left( \dot{v}^n + \left(2\omega_{ie}^n + \omega_{en}^n\right) \times v^n - g^n \right) d\tau \quad (14b)$$

Here, the subscript $\Delta v$ is used to denote the velocity increment over time interval $[t_m, t]$ and to sign difference with the velocity denotation $v$ used in (11). Although the interval of the interleaved integration is set to be the GPS-sample time in [10], it can be extended to a more general form with arbitrary length. We can determine the length of the interleaved integration interval based on the consideration of the trade-off between the convergent speed and the attenuating the negative effect of accumulative errors. That is to say, the time interval should be as long as possible to guarantee the convergent speed of the OBA method on one hand, while on the other hand, the time interval should be as short as possible to attenuate the negative effect of accumulative errors. The interleaved integrating procedure can also guarantee the numbers of the constructed vector observations without loss of intermediate GPS data samples, which is just the motivation of the derivation of the interleaved integration in [10].

In [10], there is no explicit integration formula to calculate the integration in (14). However, the velocity integration formula developed in [8, 15] can be readily applied as

$$\alpha_{\Delta v} = \sum_{k=m}^{M-1} C_{b(t_k)}^{b(0)} \int_{t_k}^{t_{k+1}} C_{b(t)}^{b(t_k)} f^b(t) dt \tag{15a}$$

$$\beta_{\Delta v} = C_{n(\tau)}^{n(0)} v^n \bigg|_{t_m}^{t} + \sum_{k=m}^{M-1} C_{n(t_k)}^{n(0)} \int_{t_k}^{t_{k+1}} C_{n(t)}^{n(t_k)} \omega_{ie}^n \times v^n dt \\ - \sum_{k=m}^{M-1} C_{n(t_k)}^{n(0)} \int_{t_k}^{t_{k+1}} C_{n(t)}^{n(t_k)} g^n(t) dt \tag{15b}$$

where $t_m = m\Delta t$. The illustration of the interleaved integration procedure is shown in Fig. 2.

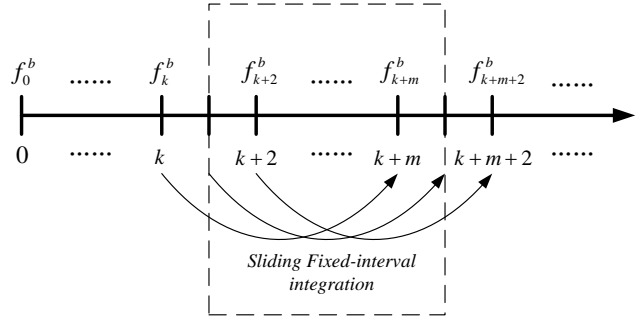

Fig. 2. Integration procedure in (14)

It should be noted that all the existing OBA methods are all unable to estimate anything other than the attitude. Since for the low-cost SINS the sensors biases have much negative effect on the attitude calculation, it is desired to extend existing OBA methods and make them be able to estimate the sensors biases coupled with the attitude for the initial alignment problem. With this consideration, a dynamic OBA method is developed in this paper, which is used to estimate the attitude at current time rather than the constant attitude at the very start of the initial alignment.

III. DYNAMIC OBA METHOD

*A. Alternative Perspective on Interleaved Integration*

Before getting into the development of the dynamic OBA method, we first give an alternative perspective on the interleaved integrating procedure, which will facilitate the derivation of the developed dynamic OBA method.

Specifically, according to the chain rule of the attitude matrix, we have

$$\begin{aligned} C_{b(t)}^{b(0)} &= C_{b(t_m)}^{b(0)} C_{b(t)}^{b(t_m)} \\ C_{n(t)}^{n(0)} &= C_{n(t_m)}^{n(0)} C_{n(t)}^{n(t_m)} \end{aligned} \tag{16}$$

Substituting (16) into (7) and integrating the resulting equation over time interval $[t_m, t]$ yield

$$\begin{aligned} & C_b^n(0) \int_{t_m}^{t} C_{b(t_m)}^{b(0)} C_{b(\tau)}^{b(t_m)} f^b d\tau \\ & = \int_{t_m}^{t} C_{n(t_m)}^{n(0)} C_{n(\tau)}^{n(t_m)} \left( \dot{v}^n + \left( 2\omega_{ie}^n + \omega_{en}^n \right) \times v^n - g^n \right) d\tau \end{aligned} \tag{17}$$

For the integrating operator $d\tau$, the time instant $t_m$ is constant and so as the corresponding attitude matrices $C_{b(t_m)}^{b(0)}$ and $C_{n(t_m)}^{n(0)}$. To this respect, (17) can be rewritten as

$$\begin{aligned} & C_b^n(0) C_{b(t_m)}^{b(0)} \int_{t_m}^{t} C_{b(\tau)}^{b(t_m)} f^b d\tau \\ & = C_{n(t_m)}^{n(0)} \int_{t_m}^{t} C_{n(\tau)}^{n(t_m)} \left( \dot{v}^n + \left( 2\omega_{ie}^n + \omega_{en}^n \right) \times v^n - g^n \right) d\tau \end{aligned} \tag{18}$$

Multiplying $C_{n(0)}^{n(t_m)}$ on both sides and incorporating the corresponding attitude matrices yield

$$\begin{aligned} & C_b^n(t_m) \int_{t_m}^{t} C_{b(\tau)}^{b(t_m)} f^b d\tau \\ & = \int_{t_m}^{t} C_{n(\tau)}^{n(t_m)} \left( \dot{v}^n + \left( 2\omega_{ie}^n + \omega_{en}^n \right) \times v^n - g^n \right) d\tau \end{aligned} \tag{19}$$

where

$$C_b^n(t_m) = C_{n(0)}^{n(t_m)} C_b^n(0) C_{b(t_m)}^{b(0)} \tag{20}$$

Eq. (19) can also be written in the form of attitude determination problem as

$$C_b^n(t_m)\alpha_{\Delta\tilde{v}} = \beta_{\Delta\tilde{v}} \tag{21}$$

with

$$\alpha_{\Delta\tilde{v}} = \int_{t_m}^{t} C_{b(\tau)}^{b(t_m)} f^b d\tau \tag{22a}$$

$$\beta_{\Delta\tilde{v}} = \int_{t_m}^{t} C_{n(\tau)}^{n(t_m)} \left( \dot{v}^n + (2\omega_{ie}^n + \omega_{en}^n) \times v^n + g^n \right) d\tau \tag{22b}$$

The relationship between (14) and (22) is given as

$$\begin{aligned}\alpha_{\Delta v} &= C_{b(t_m)}^{b(0)} \alpha_{\Delta\tilde{v}} \\ \beta_{\Delta v} &= C_{n(t_m)}^{n(0)} \beta_{\Delta\tilde{v}}\end{aligned} \tag{23}$$

Regarding (21), $C_b^n(t_m)$ can be viewed as a "dynamic constant" matrix. Here, the term "constant" in "dynamic constant" means that the attitude matrix $C_b^n(t_m)$ at time instant $t_m$ is regarded as a constant for the vector observations in following time interval $[t_m, t]$. In this respect, the determination of $C_b^n(t_m)$ based on (21) shares the same procedure as that of $C_b^n(0)$ based on (10) and (13). On the other hand, the term "dynamic" in "dynamic constant" means that the attitude matrix $C_b^n(t_m)$ is variational on the whole time interval of the initial alignment as the time instant $t_m$ is changed from 0 to $t$. The illustration of the integration procedure in (22) is shown in Fig. 3. All the attitude matrices at the marked time instant are just the aforementioned "dynamic constant" matrices.

The navigation and body frames at time instant $t_m$, i.e. $n(t_m)$ and $b(t_m)$, can also be viewed as inertial frames for the corresponding frames at the following time interval $[t_m, t]$ just the same as the $n(0)$ and $b(0)$ for the corresponding frames at time interval $[0, t]$. Since the time instant $t_m$ is changed from 0 to $t$, the corresponding frames $n(t_m)$ and $b(t_m)$ are only the "temporary inertial" frames. In the following section, the concept of "temporary inertial" frames will also be used to derive the dynamic OBA method.

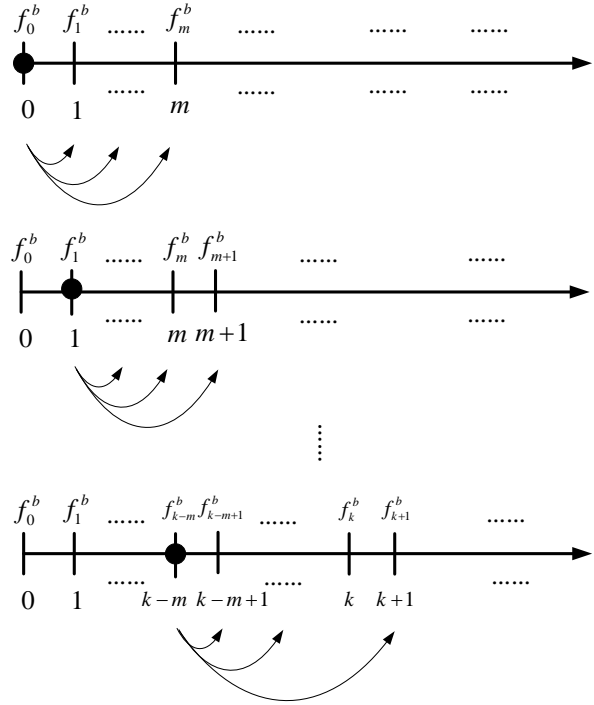

Fig. 3. Integration procedure in (22)

After the "dynamic constant" matrix $C_b^n(t_m)$ being derived, the attitude matrix at the current time instant can be readily obtained as

$$C_b^n(t) = C_{n(t_m)}^{n(t)} C_b^n(t_m) C_{b(t)}^{b(t_m)} \qquad (24)$$

where

$$\begin{aligned} C_{b(t)}^{b(t_m)} &= C_{b(0)}^{b(t_m)} C_{b(t)}^{b(0)} \\ C_{n(t_m)}^{n(t)} &= C_{n(0)}^{n(t)} C_{n(t_m)}^{n(0)} \end{aligned} \qquad (25)$$

All the attitude matrices at the right hand of (25) can be calculated according to (4). It should be noted that the attitude determination procedure (21)-(25) is not a new OBA method and it is the same as that of the interleaved integration based method.

*B. Dynamic Attitude Estimation Model*

In this section part, we will develop the dynamic OBA method with consideration of estimating the sensors biases. The derivation of such method is based on the aforementioned concepts "dynamic constant" matrix and "temporary inertial" frames. For the alternative perspective on the interleaved integration, the frames at the time instant $t_m$ are treated as inertial for the corresponding frames at the following time interval $[t_m, t]$, while in this section, the frames at time instant $t$ are treated as inertial for the corresponding frames at the previous time interval $[t_m, t]$.

Specifically for any time instant $\tau \in [t_m, t]$, according to the chain rule of the attitude matrix, we have

$$C_b^n(\tau) = C_{n(t)}^{n(\tau)} C_b^n(t) C_{b(\tau)}^{b(t)} \tag{26}$$

Substituting (26) into (5) yields

$$\dot{v}^n = C_{n(t)}^{n(\tau)} C_b^n(t) C_{b(\tau)}^{b(t)} f^b - (2\omega_{ie}^n + \omega_{en}^n) \times v^n + g^n \tag{27}$$

It should be noted that all the quantities above are functions of time $\tau$ and their time dependences on are omitted for brevity. Multiplying $C_{n(\tau)}^{n(t)}$ on both sides of (27) and reorganizing the resulting equation yield

$$C_b^n(t) C_{b(\tau)}^{b(t)} f^b = C_{n(\tau)}^{n(t)} \left( \dot{v}^n + (2\omega_{ie}^n + \omega_{en}^n) \times v^n - g^n \right) \tag{28}$$

Integrating (28) on both sides over the time interval $[t_m, t]$

$$\int_{t_m}^{t} C_b^n(t) C_{b(\tau)}^{b(t)} f^b d\tau$$
$$= \int_{t_m}^{t} C_{n(\tau)}^{n(t)} \left( \dot{v}^n + (2\omega_{ie}^n + \omega_{en}^n) \times v^n - g^n \right) d\tau \tag{29}$$

For the integrating operator $d\tau$, the time instant $t$ is constant and so as the corresponding $C_b^n(t)$. To this respect, (29) can be rewritten as

$$C_b^n(t) \alpha_{dv} = \beta_{dv} \tag{30}$$

with

$$\alpha_{dv} = \int_{t_m}^{t} C_{b(\tau)}^{b(t)} f^b d\tau \tag{31a}$$

$$\beta_{dv} = \int_{t_m}^{t} C_{n(\tau)}^{n(t)} \left( \dot{v}^n + (2\omega_{ie}^n + \omega_{en}^n) \times v^n - g^n \right) d\tau \tag{31b}$$

It is shown that we have made use of the similar "temporary inertial" concept in going from (17) to (18) and (29) to (30). The reference frame $C_b^n(t)$ corresponding to the "temporary inertial" frames can also be viewed as a "dynamic constant" matrix. In last section, the constant characteristic of the "dynamic constant" matrix $C_b^n(t_m)$ is used for the static attitude determination. In contrast, the dynamic characteristic of the "dynamic constant" matrix $C_b^n(t)$ will be used for the following dynamic attitude estimation. When the attitude matrix $C_b^n(t)$ is treated as a dynamic value, the gyroscopes biases can be easily incorporated into the dynamic attitude update model. Considering the contaminated bias and noise, the direct output of the gyroscopes $\tilde{\omega}_{ib}^b(t)$ is modeled as

$$\begin{aligned}\tilde{\omega}_{ib}^b(t) &= \omega_{ib}^b(t) + \varepsilon^b(t) + \eta_{gv} \\ \dot{\varepsilon}^b(t) &= \eta_{gu}\end{aligned} \quad (32)$$

where $\varepsilon^b$ is the gyro drift derived from a random walk process. $\eta_{gv}$ and $\eta_{gu}$ are independent zero-mean Gaussian white-noise processes with spectral densities.

Reorganizing (32) and substituting the result into (1) give the following attitude model

$$\dot{C}_b^n(t) = C_b^n(t)\left(\tilde{\omega}_{ib}^b(t) - \varepsilon^b(t) - C_b^n(t)^T \omega_{in}^n(t) - \eta_{gv}\right)\times \quad (33)$$

Eq. (33) and (30) constitute the typical dynamic attitude estimation model with (33) being the process model and (30) the measurement model. With the developed dynamic model, many existing dynamic attitude estimation methods known as filtering methods can be readily applied. Before getting involved in explicit details of filtering algorithms that will be presented in next section, we first extend the velocity integration formula developed in [8, 15] to solve the integration (31).

The velocity integration formula developed in [8, 15] can not be applied in (31) in its current form. Next, we will show how the integration in (31) can be calculated using the velocity integration formula.

The integration in (31a) can be expressed as

$$\int_{t_m}^{t} C_{b(\tau)}^{b(t)} f^b d\tau = \int_{t_m}^{t} C_{b(0)}^{b(t)} C_{b(\tau)}^{b(0)} f^b d\tau = C_{b(0)}^{b(t)} \int_{t_m}^{t} C_{b(\tau)}^{b(0)} f^b d\tau \tag{34}$$

As it is shown, the integration on the right hand of (34) is the same as that in (14a). Therefore, the integration in (31a) can be calculated as

$$\alpha_{dv} = C_{b(0)}^{b(t)} \sum_{k=m}^{M-1} C_{b(t_k)}^{b(0)} \int_{t_k}^{t_{k+1}} C_{b(t)}^{b(t_k)} f^b(t) dt \tag{35}$$

The first part of the integration (31b) can be developed as

$$\int_{t_m}^{t} C_{n(\tau)}^{n(t)} \dot{v}^n d\tau = C_{n(\tau)}^{n(t)} v^n \Big|_{t_m}^{t} - \int_{t_m}^{t} C_{n(\tau)}^{n(t)} \omega_{in}^n \times v^n d\tau$$
$$= v^n(t) - C_{n(t_m)}^{n(t)} v^n(t_m) - \int_{t_m}^{t} C_{n(\tau)}^{n(t)} \omega_{in}^n \times v^n d\tau \tag{36}$$

Substituting (36) into (31b) yields

$$\beta_{dv} = v^n(t) - C_{n(t_m)}^{n(t)} v^n(t_m)$$
$$+ \int_{t_m}^{t} C_{n(\tau)}^{n(t)} \omega_{ie}^n \times v^n d\tau - \int_{t_m}^{t} C_{n(\tau)}^{n(t)} g^n d\tau \tag{37}$$

Similar to the procedure in (34), the first integration on the right hand of (37) can be expressed and calculated as

$$\int_{t_m}^{t} C_{n(\tau)}^{n(t)} \omega_{ie}^n \times v^n d\tau = \int_{t_m}^{t} C_{n(0)}^{n(t)} C_{n(\tau)}^{n(0)} \omega_{ie}^n \times v^n d\tau$$
$$= C_{n(0)}^{n(t)} \sum_{k=m}^{M-1} C_{n(t_k)}^{n(0)} \int_{t_k}^{t_{k+1}} C_{n(t)}^{n(t_k)} \omega_{ie}^n \times v^n dt \tag{38}$$

The second integration on the right hand of (37) can be expressed and calculated as

$$\int_{t_m}^{t} C_{n(\tau)}^{n(t)} g^n d\tau = \int_{t_m}^{t} C_{n(0)}^{n(t)} C_{n(\tau)}^{n(0)} g^n d\tau$$
$$= C_{n(0)}^{n(t)} \sum_{k=m}^{M-1} C_{n(t_k)}^{n(0)} \int_{t_k}^{t_{k+1}} C_{n(t)}^{n(t_k)} g^n(t) dt \tag{39}$$

Substituting (38) and (39) into the right side of (37) yields

$$\beta_{dv} = v^n(t) - C_{n(t_m)}^{n(t)} v^n(t_m)$$
$$+ C_{n(0)}^{n(t)} \left( \sum_{k=m}^{M-1} C_{n(t_k)}^{n(0)} \int_{t_k}^{t_{k+1}} C_{n(t)}^{n(t_k)} \omega_{ie}^n \times v^n dt - \sum_{k=m}^{M-1} C_{n(t_k)}^{n(0)} \int_{t_k}^{t_{k+1}} C_{n(t)}^{n(t_k)} g^n(t) dt \right) \tag{40}$$

The illustration of the developed dynamic attitude determination problem and the associated vector observations (31) are shown in Fig. 4. In Fig. 4, $x_k$ is the state of the dynamic model, which includes attitude and gyroscopes biases. It should be noted that, although the vector observations are constructed based on the interleaved integration procedure, they can also be constructed based on the integration procedure in (11). Let $t_m = 0$, the integration procedure in (31) is just the same as that in (11). The properties of the two integration procedures discussed in the last section for the OBA method are also suitable for the dynamic attitude estimation problem.

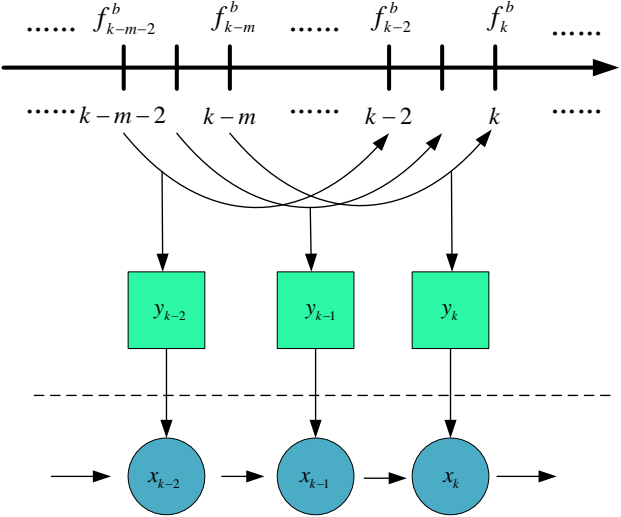

Fig. 4. Integration procedure in (31) for the dynamic model (32)

### C. Modified UnScented QUaternion Estimator

Since the attitude is usually parameterized in terms of quaternion for on-computer application, most of the existing filtering methods for attitude estimation are devoted to handle the quaternion normalization constraint in the filtering recursion, such as the multiplicative extended Kalman filter [14], the UnScented QUaternion Estimator (USQUE) [16, 17] and its modified version (MUSQUE) [18, 19]. In this paper, the MUSQUE is used. Considering clarity and completeness, the explicit filtering procedure of MUSQUE for the investigated dynamic attitude estimation problem is presented in this section part. Parameterizing the attitude in terms of quaternion, the attitude kinematics equation (33) in its discrete-time form is given by

$$q_k = ff\left(q_{k-1}, \varepsilon_{k-1}, \omega_{ib}^b, \omega_{in}^n\right) \tag{41}$$

Eq. (41) is a general form of the discrete-time attitude kinematics equation in terms of quaternion. The discrete-time attitude kinematics equation is virtually the on-computer calculation procedure of the attitude. Various fruitful algorithms have been developed for the on-computer attitude calculation, say single sample algorithm, double sample algorithm and so on. We can select the appropriate one based on the consideration of the algorithm precision and complexity. Since this paper focuses on the initial alignment method development, the explicit procedures of these calculation algorithms are not presented and only a general form as shown in (41) is presented.

The MUSQUE is virtually an extension of the unscented Kalman filter (UKF) [20, 21] with consideration of the quaternion norm constraint. Since the norm constraint of the quaternion can be easily destroyed by the weighted-mean computation inherent in the UKF, the attitude quaternion can not be used directly as the filtering state. In order to overcome this problem the MUSQUE uses the quaternion for global nonsingular attitude representation and unconstrained generalized Rodrigues parameter (GRP) for local attitude representation and filtering. Given the error quaternion $\delta q = \begin{bmatrix} \delta \rho^T & \delta q_4 \end{bmatrix}^T$, where $\delta \rho$ is the vector part of the quaternion $\delta q$, the corresponding GRP representation $\delta p$ is given by

$$\delta p = f \frac{\delta \rho}{a + \delta q_4} \tag{42}$$

where $a$ is a parameter from 0 to 1, and $f$ is a scale factor.

In the MUSQUE, the state is selected as $x_k = [\delta \rho_k ; \varepsilon_k]$. Given the state estimate $\hat{x}_{k-1}$ at time instant $k-1$, the corresponding state covariance $P_{k-1}$ and the quaternion estimate $\hat{q}_{k-1}$, the aim of MUSQUE is to determine the state estimate $\hat{x}_k$ and quaternion estimate $\hat{q}_k$ at time instant $k$. The explicit filtering procedure is given as follows

*Time Update*

Generate the sigma points as

$$\chi_{k-1} = \begin{bmatrix} \chi_{k-1}^{\delta p} \\ \chi_{k-1}^{\varepsilon} \end{bmatrix} = \hat{x}_{k-1} \pm \sqrt{(n+\kappa)P_{k-1}} \tag{43}$$

where $\kappa$ is a tune parameter which is usually set to $3-n$ in the UKF to capture some higher order information of the distribution with $n$ denoting the state dimension [19]. The $i$-th component of $\chi_{k-1}^{\delta p}$ is denoted as $\chi_{k-1}^{\delta p}(i)$ and its quaternion form is denoted as $\delta q_{k-1}(i) = \begin{bmatrix} \delta \rho_{k-1}^T(i) & \delta q_{4,k-1}(i) \end{bmatrix}^T$ which can be calculated according to the reverse form of (42) as

$$\delta q_{4,k-1}(i) = \frac{-a\left\|\chi_{k-1}^{\delta p}(i)\right\|^2 + f\sqrt{f^2 + (1-a^2)\left\|\chi_{k-1}^{\delta p}(i)\right\|^2}}{f^2 + \left\|\chi_{k-1}^{\delta p}(i)\right\|^2} \tag{44a}$$

$$\delta \rho_{k-1}(i) = f^{-1}\left[a + \delta q_{4,k-1}(i)\right]\chi_{k-1}^{\delta p}(i) \tag{44b}$$

The corresponding quaternion based sigma points $\chi_{k-1}^q(i)$ are given as

$$\chi_{k-1}^q(i) = \delta q_{k-1}(i) \otimes \hat{q}_{k-1} \tag{45}$$

Propagate the quaternion and gyroscopes bias based sigma points through the process model (41)

$$\chi_{k|k-1}^q(i) = ff\left(\chi_{k-1}^q(i), \chi_{k-1}^{\varepsilon}(i), \omega_{ib}^b, \omega_{in}^n\right) \tag{46}$$

Since the gyroscopes bias is viewed as constant, the propagated gyroscopes bias based sigma points are given as

$$\chi_{k|k-1}^{\varepsilon}(i) = \chi_{k-1}^{\varepsilon}(i) \tag{47}$$

The prediction of the attitude quaternion is given by the quaternion averaging algorithm as [22]

$$\hat{q}_{k|k-1} = \arg\max_{q \in S^3} q^T A q \tag{48}$$

where

$$A = \frac{1}{2n}\sum_{i=1}^{2n} \chi_{k|k-1}^q(i) \cdot \chi_{k|k-1}^q(i)^T \tag{49}$$

The predicted quaternion error based sigma points are given by

$$\delta q_{k|k-1}(i) = \chi^{q}_{k|k-1}(i) \otimes \left[\hat{q}_{k|k-1}\right]^{-1} \tag{50}$$

Denote $\delta \hat{q}_{k|k-1}(i) = \left[\delta \rho^{T}_{k|k-1}(i) \quad \delta q_{4,k|k-1}(i)\right]^{T}$, the corresponding predicted GRP based sigma points are given by

$$\chi^{\delta p}_{k|k-1}(i) = f \frac{\delta \rho_{k|k-1}(i)}{a + \delta q_{4,k|k-1}(i)} \tag{51}$$

Denote $\chi_{k|k-1} = \left[\chi^{\delta pT}_{k|k-1} \quad \chi^{\varepsilon T}_{k|k-1}\right]^{T}$, the predicted mean and covariance of the state are given by

$$\hat{x}_{k|k-1} = \frac{1}{2n} \sum_{i=1}^{2n} \chi_{k|k-1}(i) \tag{52a}$$

$$P_{k|k-1} = \frac{1}{2n} \sum_{i=1}^{2n} \left[\chi_{k|k-1}(i) - \hat{x}_{k|k-1}\right]\left[\chi_{k|k-1}(i) - \hat{x}_{k|k-1}\right]^{T} + Q_{k-1} \tag{52b}$$

where $Q_{k-1}$ is the covariance of the process noise.

*Measurement Update*

Regenerate the sigma points as

$$\chi^{*}_{k|k-1} = \begin{bmatrix} \chi^{*\delta p}_{k|k-1} \\ \chi^{*\varepsilon}_{k|k-1} \end{bmatrix} = \hat{x}_{k|k-1} \pm \sqrt{(n+\kappa)P_{k|k-1}} \tag{53}$$

Denote the quaternion form of $\chi^{*\delta p}_{k|k-1}(i)$ as

$$\delta q^{*}_{k|k-1}(i) = \left[\delta \rho^{*T}_{k|k-1}(i) \quad \delta q^{*}_{4,k|k-1}(i)\right]^{T}$$

which can be calculated according to the reverse form of (42) as

$$\delta q^{*}_{4,k|k-1}(i) = \frac{-a\left\|\chi^{*\delta p}_{k|k-1}(i)\right\|^{2} + f\sqrt{f^{2} + (1-a^{2})\left\|\chi^{*\delta p}_{k|k-1}(i)\right\|^{2}}}{f^{2} + \left\|\chi^{*\delta p}_{k|k-1}(i)\right\|^{2}} \tag{54a}$$

$$\delta \rho^{*}_{k|k-1}(i) = f^{-1}\left[a + \delta q^{*}_{4,k|k-1}(i)\right]\chi^{*\delta p}_{k|k-1}(i) \tag{54b}$$

The quaternion based sigma points corresponding to $\delta q^*_{k|k-1}(i)$ is given by

$$\chi^{*q}_{k|k-1}(i) = \delta q^*_{k|k-1}(i) \otimes \hat{q}_{k|k-1} \tag{55}$$

Propagate the quaternion based sigma points through the measurement model (30) as

$$\chi^{\beta}_k(i) = C\left(\chi^{*q}_{k|k-1}(i)\right) \alpha_{dv,k} \tag{56}$$

where $C\left(\chi^{*q}_{k|k-1}(i)\right)$ denotes the attitude matrix corresponding the quaternion $\chi^{*q}_{k|k-1}(i)$.

The mean and covariance of the measurement are given by

$$\hat{y}_k = \frac{1}{2n} \sum_{i=1}^{2n} \chi^{\beta}_k(i) \tag{57a}$$

$$P_{y,k} = \frac{1}{2n} \sum_{i=1}^{2n} \left[\chi^{\beta}_k(i) - \hat{y}_k\right]\left[\chi^{\beta}_k(i) - \hat{y}_k\right]^T + R_k \tag{57b}$$

where $R_k$ is the covariance of the measurement noise. The cross-correlation covariance of the state and measurement is given by

$$P_{xy,k} = \frac{1}{2n} \sum_{i=1}^{2n} \left[\chi_{k|k-1}(i) - \hat{x}_{k|k-1}\right]\left[\chi^{\beta}_k(i) - \hat{y}_k\right]^T \tag{57c}$$

The state estimate and corresponding covariance are given by

$$\hat{x}_k = \hat{x}_{k|k-1} + K_k\left(\beta_{dv,k} - \hat{y}_k\right) \tag{58a}$$

$$P_k = P_{k|k-1} - K_k P_{y,k} K_k^T \tag{58b}$$

where

$$K_k = P_{xy,k}\left(P_{y,k}\right)^{-1} \tag{58c}$$

*Attitude Update*

Denote $\hat{x}_k = \begin{bmatrix} \delta\hat{p}_k^T & \hat{\varepsilon}_k^T \end{bmatrix}^T$. The quaternion corresponding to $\delta\hat{p}_k$ is denoted as $\delta\hat{q}_k = \begin{bmatrix} \delta\rho_k^T & \delta q_{4,k} \end{bmatrix}^T$ which can be calculated as

$$\delta q_{4,k} = \frac{-a\|\delta \hat{p}_k\|^2 + f\sqrt{f^2 + (1-a^2)\|\delta \hat{p}_k\|^2}}{f^2 + \|\delta \hat{p}_k\|^2} \quad (59a)$$

$$\delta \rho_k = f^{-1}\left[a + \delta q_{4,k}(i)\right]\delta \hat{p}_k \quad (59b)$$

The quaternion estimate is given by

$$\hat{q}_k = \delta \hat{q}_k \otimes \hat{q}_{k|k-1} \quad (60)$$

Reset the first three elements (GRP) of $\hat{x}_k$ to zero and go to the next filtering cycle.

## IV. PERFORMANCE EVALUATION THROUGH FIELD TEST

In this section, the vector observations construction procedures for both the static and dynamic OBA methods are evaluated using field test data. Specifically, the following four initial alignment methods are evaluated:

*q method: full integration*

*q method: partial integration*

*filter: full integration*

*filter: partial integration*

The term "full integration" means the integration procedure in (11) and "partial integration" the interleaved integration procedure in (14) or (31). The term "q method" is corresponding to the OBA method as the Davenport's q method is used to calculate the constant attitude matrix. The term "filter" is corresponding to the dynamic OBA method as the MUSQUE algorithm is used to calculate the dynamic attitude matrix. In the partial integration procedure, 100 interleaved samples are used, i.e. $t - t_m = 100\Delta t$ for (14) and (31).

As is discussed in the last section, the two different integration procedures mainly affect the convergent speed and steady-state estimate of the corresponding attitude determination methods. In this section, we will evaluate the two characteristics of the aforementioned four initial alignment methods using car-mounted field test data collected on SINS with different levels of precision. One is the navigation-grade SINS equipped with a triad of ring laser gyroscopes and accelerometers. The other is the low-grade SINS equipped with a triad of MEMS gyroscopes and accelerometers. The specifications of the navigation-grade and low-grade inertial sensors are listed in Tables I and II, respectively.

TABLE I.  SPECIFICATIONS OF THE NAVIGATION-GRADE INERTIAL SENSORS

|  | Gyro | Accelerometer |
| --- | --- | --- |
| Dynamic | $\pm 400°/s$ | $\pm 3g$ |
| Update rate | 100Hz | 100Hz |
| Bias | $\leq 0.05°/h$ | $\leq 50\mu g$ |
| Bias stability | $\leq 0.01°/h$ | $\leq 50\mu g$ |

TABLE II.  SPECIFICATIONS OF THE LOW-GRADE INERTIAL SENSORS

|  | Gyro | Accelerometer |
| --- | --- | --- |
| Dynamic | $\pm 150°/s$ | $\pm 10g$ |
| Update rate | 100Hz | 100Hz |
| Bias | $\leq 0.5°/s$ | $\leq 0.005g$ |
| Bias stability | $\leq 0.02°/s$ | $\leq 0.001g$ |

There is also a GPS antenna installed outside the cabin on the top of the car and the SINS/GPS integrated navigation result is used as the reference data to compare with. The raw measurements (1 *Hz*) of GPS were linearly interpolated to obtain the velocity and position at both ends of the update interval.

A.  *Testing Results Using the Navigation-grade SINS*

According to [8], the OBA method can reach a satisfactory performance for the navigation-grade SINS after only 100s, therefore, we also make use of 100s collected data to test the involved initial alignment methods in this paper. The testing results of the four initial alignment methods using the navigation-grade SINS data are presented in Fig. 5-10, respectively. Fig. 5 and 6 plot the pitch angles estimates and estimate error, Fig. 7 and 8 the roll angles, Fig. 9 and 10 the yaw angles, respectively. The stead-state attitude error summary of the four initial alignment methods is listed in Table III for comparison. It is shown that the full integration procedure can reach a much faster convergent speed and better stead-state estimate performance than the partial integration procedure. This is because that in the full integration procedure the sampling data collected from the inertial sensors is used more times and, in turn, more adequately than that in the partial integration procedure. The improvement in terms of convergent speed and steady-state precision by the full integration procedure is guaranteed by the high-grade specifications of the navigation-grade inertial sensors. That is to say, making use of the sampling data collected from the navigation-grade inertial sensors repeatedly will not cause accumulative errors. Therefore, for the OBA of the high-grade SINS, the full integration procedure will be more celebrated. Moreover, with the full integration procedure the dynamic method developed in this paper is a little better than the static method. However, the computational cost of the dynamic method is much larger than that of the static method. All the initial alignment methods are implemented using Matlab on a computer with a 2.66G CPU, 2.0G memory and the Windows7 operating system. In our experiment, the cost time of processing the $100s$ data is about $52s$ for dynamic OBA method with full integration and $8s$ for static OBA method with full integration. If the OBA methods are only regarded as coarse alignment methods which are used to provide rough attitude for the following fine alignment, the static OBA method will be more celebrated due to its less computational cost.

The proposed dynamic OBA method shows no much improvement over the traditional method is mainly due to the inability of the proposed method to estimate the accelerometer biases. The accelerometer biases mainly determine the ultimate precision of the initial alignment. If the negative effect of the gyroscopes biases is not so obvious as for the navigation-grade SINS, the steady-state alignment precision is mainly determined by the accelerometer biases. Since both the static and dynamic OBA methods are unable to estimate the accelerometer biases, their performance is similar to each other in this case. Advanced method that is able to estimate the accelerometer biases can be expected to further improve the alignment performance of the navigation-grade SINS.

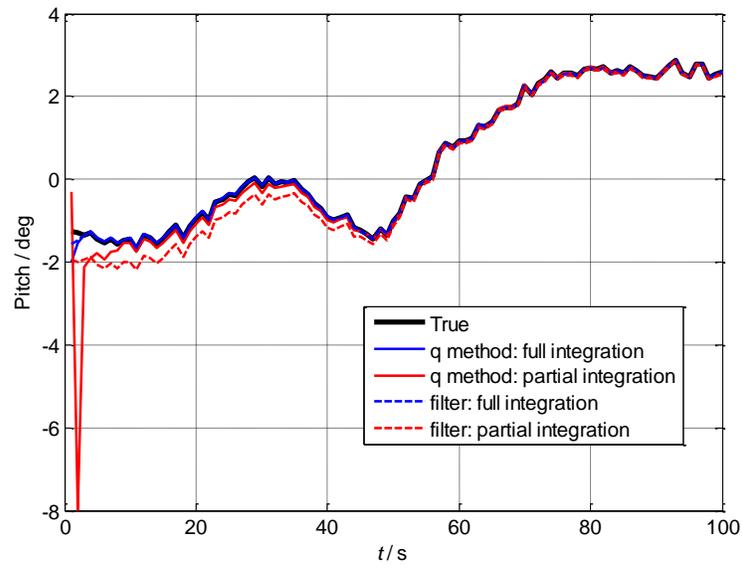

Fig. 5. Pitch angles estimates

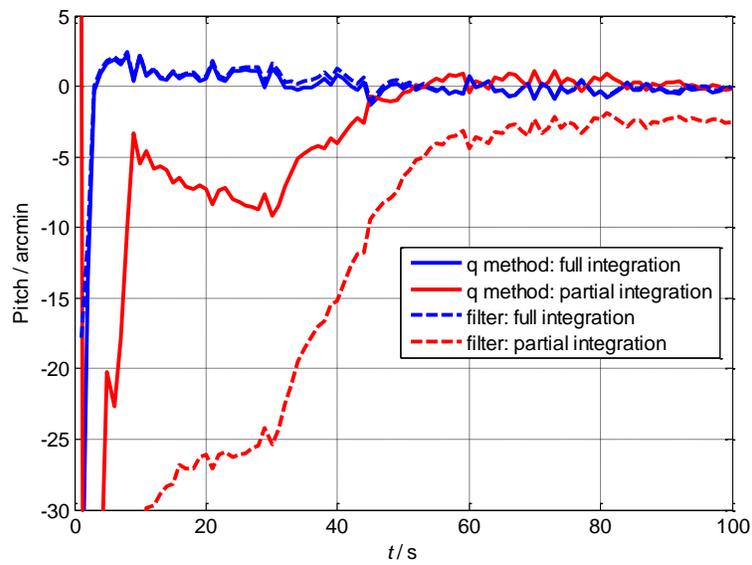

Fig. 6. Pitch angles estimate error

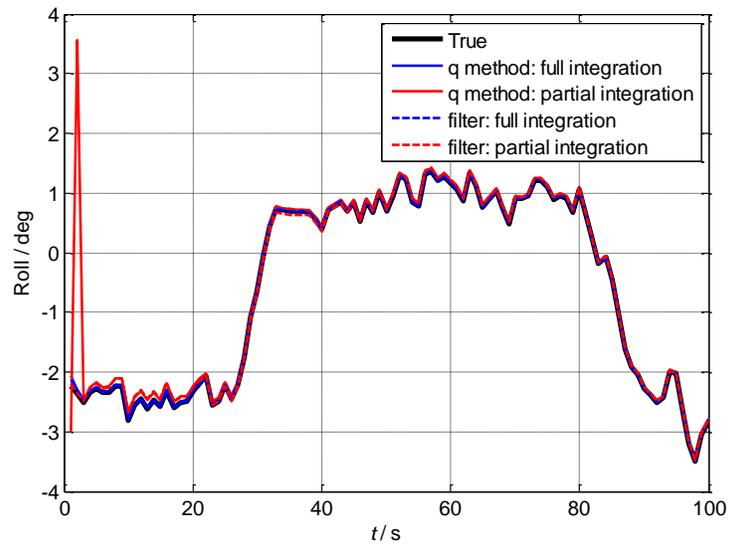

Fig. 7. Roll angles estimates

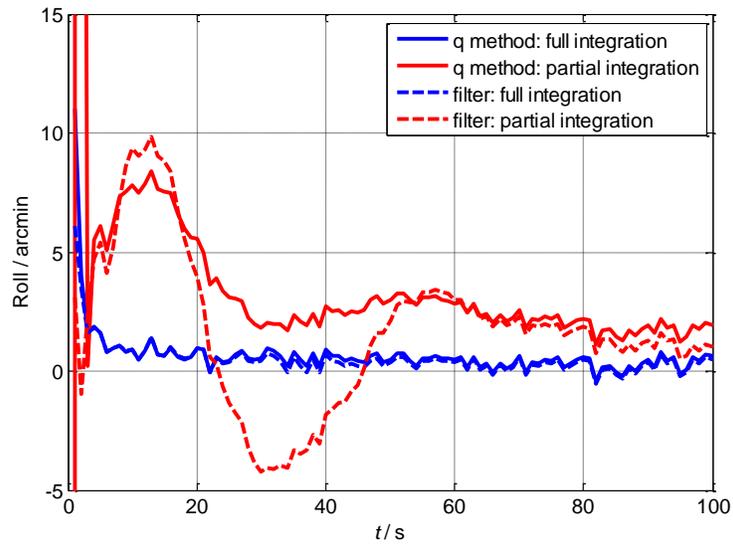

Fig. 8. Roll angles estimate error

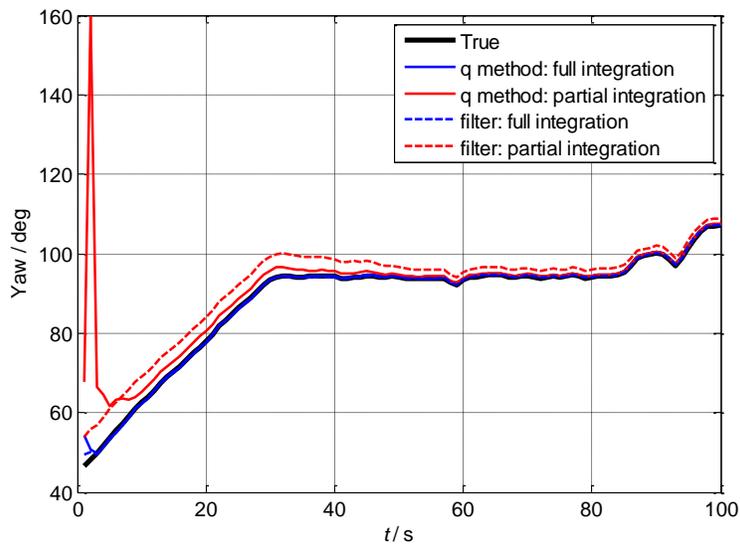

Fig. 9. Yaw angles estimates

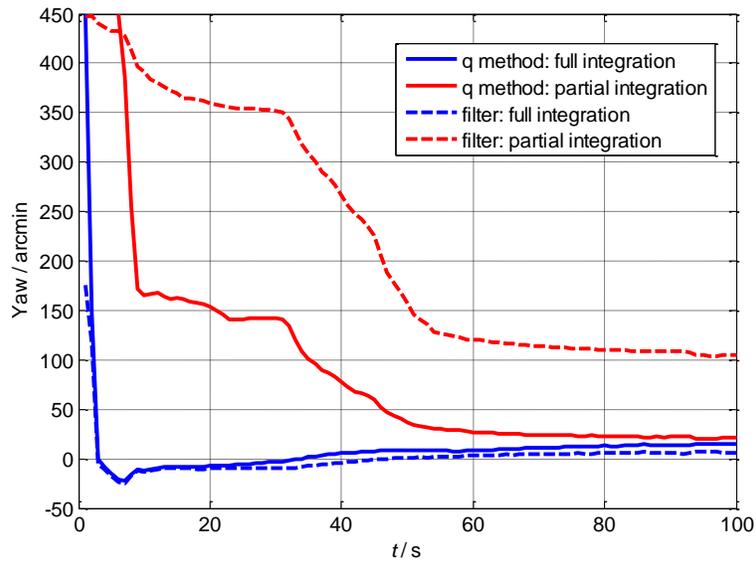

Fig. 10. Yaw angles estimate error

Table III. ATTITUDE ERROR SUMMARY USING NAVIGATION-GRADE INERTIAL SENSORS (unit: arcmin)

|  | Pitch | Roll | Yaw |
| --- | --- | --- | --- |
| q method: full integration | -0.18 | 0.62 | 14.41 |
| q method: partial integration | -0.23 | 1.92 | 21.15 |
| filter: full integration | -0.07 | 0.48 | 5.88 |
| filter: partial integration | -2.51 | 0.99 | 104.4 |

The main advantage of the developed dynamic method is its ability to estimate the gyroscopes biases coupled with the attitude. The gyroscopes biases estimate results with different integration procedures are presented in Fig. 11 and 12, respectively. Meanwhile, the gyroscopes biases estimate results using the SINS/GPS integration are also plotted in Fig. 11 and 12 to compare with. For the SINS/GPS integration, the initial gyroscopes biases are assumed to be zero, which is the same as that in the dynamic OBA methods. However, all the attitude, velocity and position are properly initialized. That is to say, the SINS/GPS integration is actually carried out after the attitude alignment. As it is shown, the gyroscopes biases estimate results using the full integration procedure are more close to the results of the SINS/GPS integration than that of the partial integration procedure. However, the results of all the three estimate procedure are not so well. This is because that for the high-grade SINS, the observability of the biases errors is feeble and much longer time is needed to converge the estimate results [6, 7].

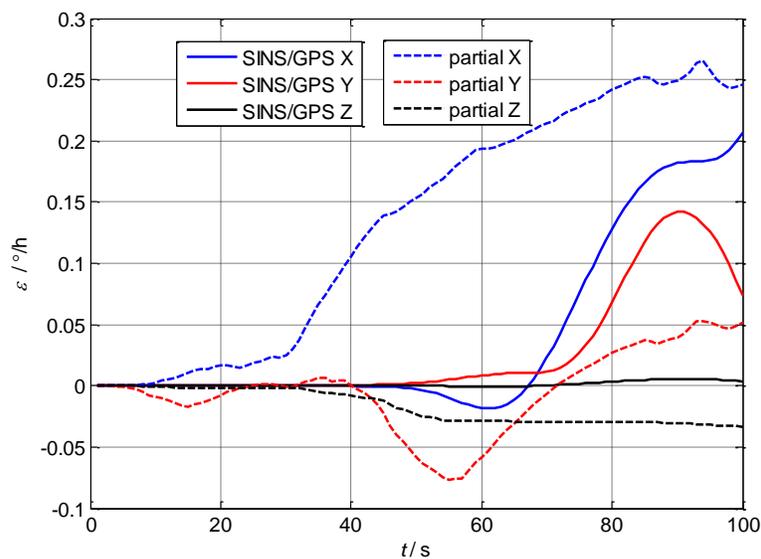

Fig. 11. Gyroscopes biases estimates of the "*filter: partial integration*" and SINS/GPS

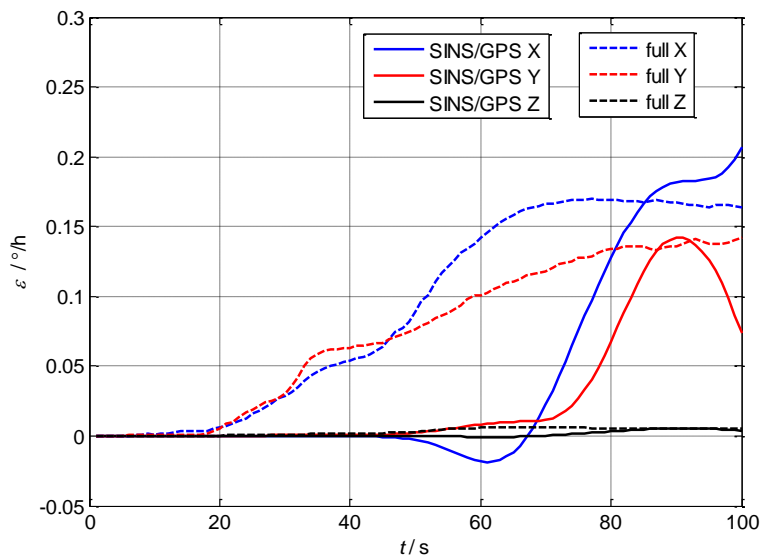

Fig. 12. Gyroscopes biases estimates of the "*filter:full integration*" and SINS/GPS

B.  Testing Results Using the Low-grade SINS

In this test, the car with low-grade SINS was moving in a large circle around the sports grounds with a good sky view of the GPS antennas. Two circles of test data with different velocities are collected to evaluate the alignment methods. In the first circle, 250s were taken for the car to make one circuit around the sports grounds. The testing results of the four initial alignment methods using the first data segment are presented in Fig. 13-18, respectively. Fig. 13 and 14 plot the pitch angles estimate and estimate error, Fig. 15 and 16 the roll angles, Fig. 17 and 18 the yaw angles, respectively. The stead-state attitude error summary of the four initial alignment methods is listed in Table IV for comparison. It is shown that both the two static OBA methods are no longer applicable. This is because that the gyroscopes biases have a much negative effect on the accuracy of the attitude and the static OBA methods are unable to estimate the biases during the attitude alignment. For the dynamic OBA methods developed in this paper, the partial integration procedure is effective while the full integration procedure is noneffective just as the static methods. This is because that the inertial sensors biases will cumulate in the constructed vector observations and when the integrating interval increases with time just as the full integration procedure the cumulated errors will be larger and larger, resulting in much degraded vector observations and, in turn, much degraded attitude estimate. The effectiveness of the dynamic OBA method with partial integration procedure lies in its ability to estimate the gyroscopes biases. The gyroscopes biases estimate results with partial integration procedure along with the results of the SINS/GPS integration are presented in Fig. 19. It is shown that biases estimate results of the dynamic OBA method with partial integration procedure are comparable with that of the SINS/GPS integration. For comparison, the gyroscopes biases estimate results with full integration procedure along with the results of the SINS/GPS integration are presented in Fig. 20. It is shown that biases estimate results of the dynamic OBA method with full integration procedure are much degraded. Therefore, the dynamic OBA method with partial integration procedure is most celebrated for the low-grade SINS.

It is shown that the estimate of the gyroscopes bias in the vertical direction has a much slower convergence speed than that of the gyroscopes biases. This is mainly due to the different degrees of observability. For more thorough details about the observability of the initial alignment, the reader is referred to the systematical and comprehensive study [7] for this problem. Although it appears that the test time is a little short to obtain a clear result for the vertical gyroscopes bias, it does not affect the demonstration of the validity of the proposed dynamic OBA method. This is because that the performance of the proposed dynamic OBA method in estimating the gyroscopes biases is comparable with that of the SINS/GPS integration which is the accredited method in estimating the sensor biases. However, it should be noted that the SINS/GPS integration is founded on the linearized system error models and necessitates the initial alignment to provide roughly known initial attitude to guarantee the validity of such linearization. If the attitude can not be aligned appropriately, the performance of the SINS/GPS integration will be much degraded. Therefore, the investigated initial alignment method is still necessary in practical application. It can be seen in Fig. 13-18 and Tab. IV that the dynamic OBA method with partial integration procedure can align the attitude quite well. Therefore, the subsequent stage of SINS/GPS integration can be readily carried out after the dynamic OBA. Meanwhile, the attitude alignment time should not be too long due to some special military requirements in terms of maneuverability. In this respect, the test time is appropriate for the attitude alignment. The calibration of the sensor biases can be further carried out in the subsequent SINS/GPS integration stage.

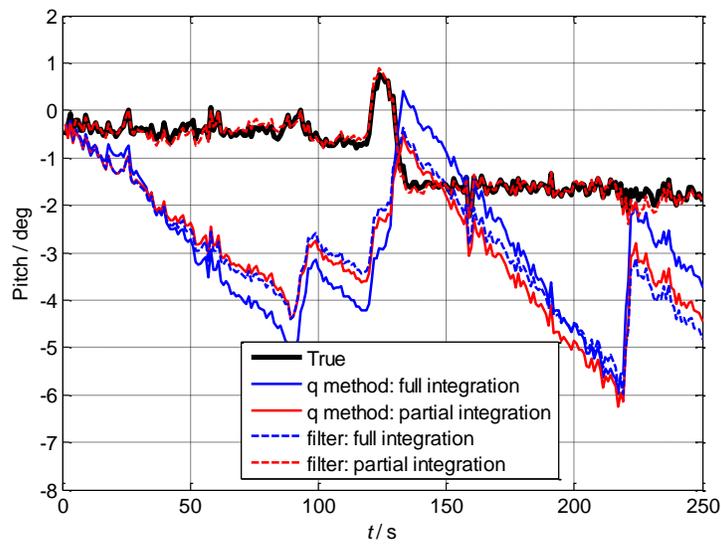

Fig. 13. Pitch angles estimates (first circle)

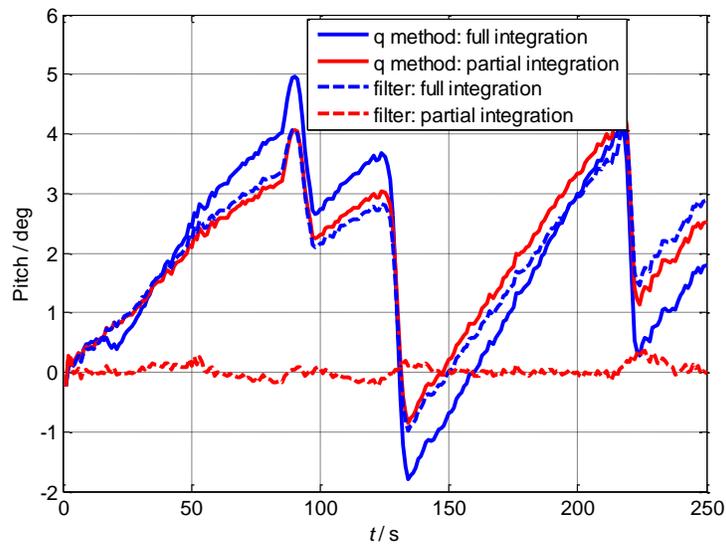

Fig. 14. Pitch angles estimate error (first circle)

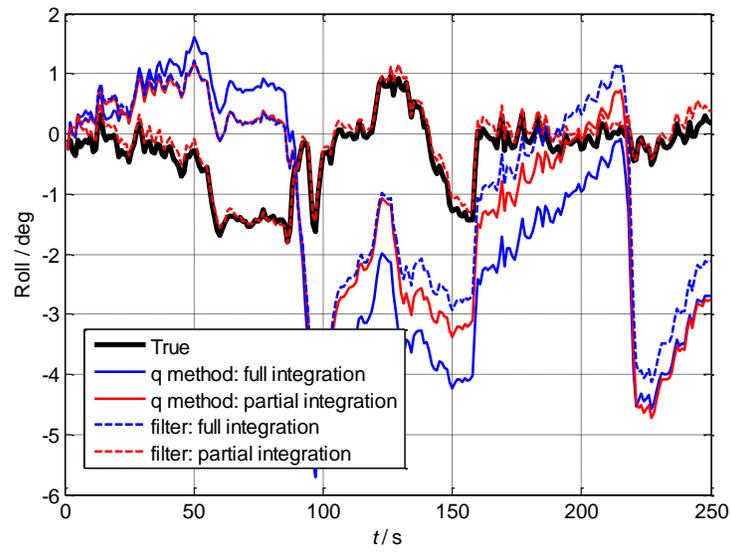

Fig. 15. Roll angles estimates (first circle)

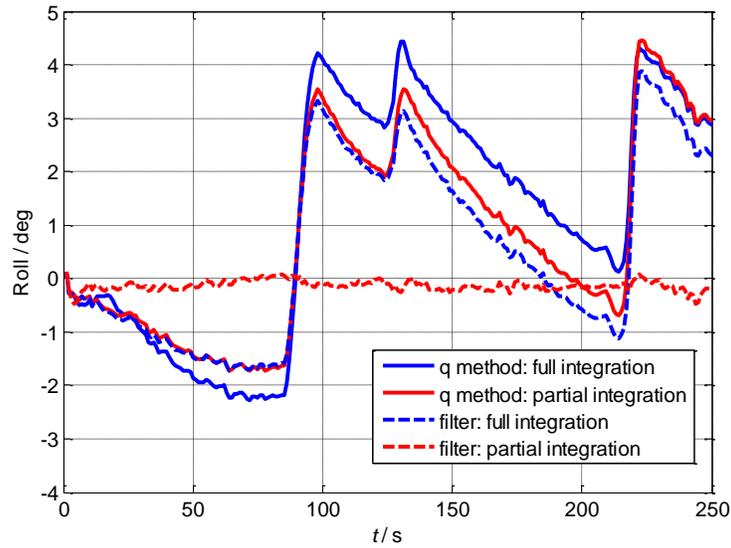

Fig. 16. Roll angles estimate error (first circle)

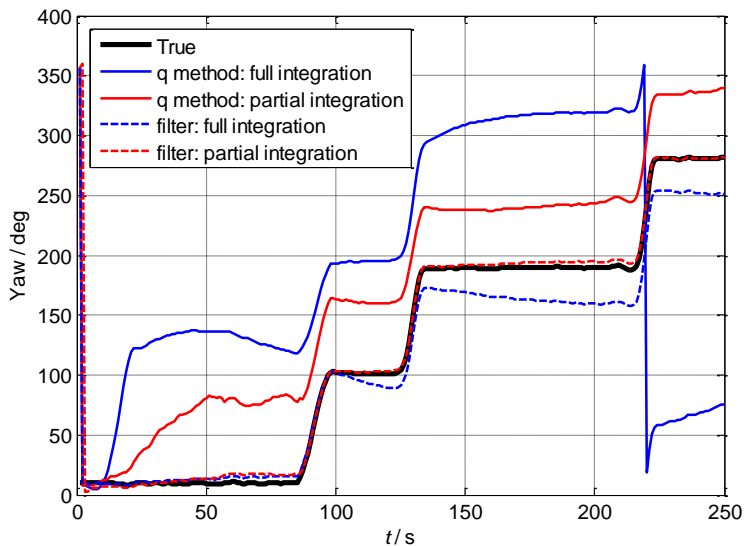

Fig. 17. Yaw angles estimates (first circle)

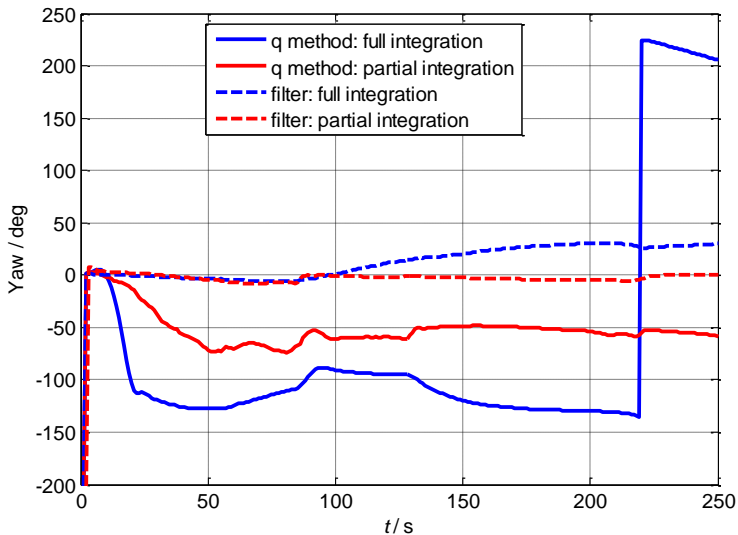

Fig. 18. Yaw angles estimate error (first circle)

Table IV. ATTITUDE ERROR SUMMARY USING LOW-GRADE INERTIAL SENSORS (first circle, unit: deg)

|  | Pitch | Roll | Yaw |
| --- | --- | --- | --- |
| q method: full integration | 1.77 | 2.97 | 207.81 |
| q method: partial | 2.42 | 2.99 | -57.58 |
| filter: full integration | 2.87 | 2.35 | 29.69 |

| filter: partial integration | 0.11 | -0.18 | -0.24 |

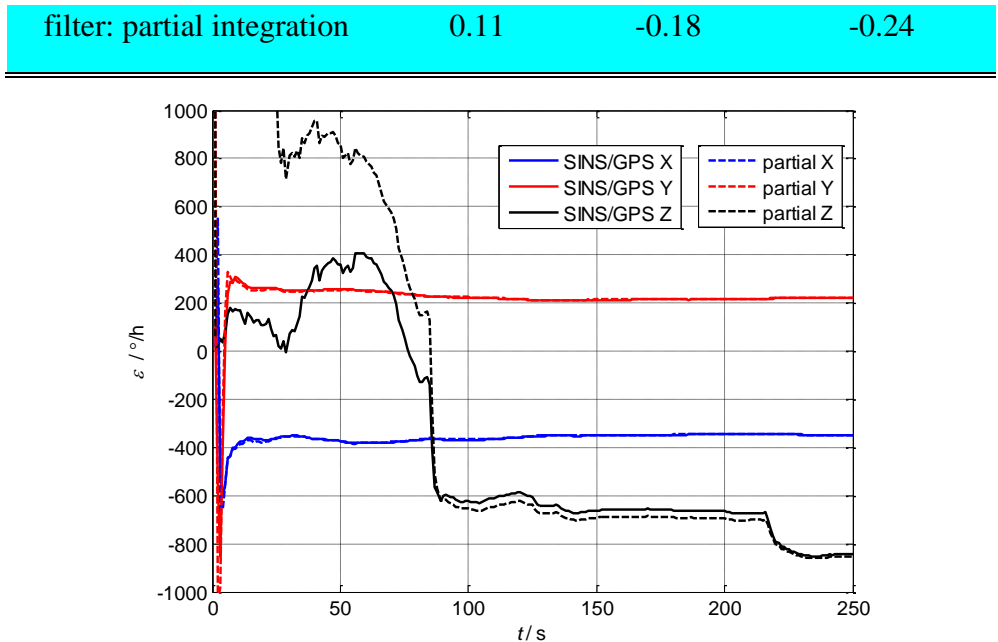

Fig. 19. Gyroscopes biases estimates of the "*filter: partial integration*" and SINS/GPS (first circle)

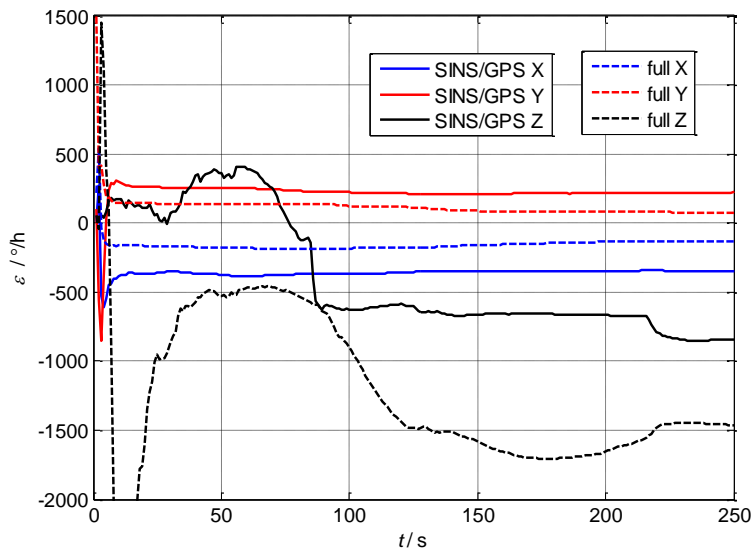

Fig. 20. Gyroscopes biases estimates of the "*filter: full integration*" and SINS/GPS (first circle)

In the second circle, 150s were taken for the car to make one circuit around the sports grounds. The testing results of the four initial alignment methods using the second data segment are presented in Fig. 21-26, respectively. Fig. 21 and 22 plot the pitch angles estimate and estimate error, Fig. 23 and 24 the roll angles, Fig. 25 and 26 the yaw angles, respectively. The stead-state attitude error summary of the four initial alignment methods is listed in Table V for comparison. The gyroscopes biases estimate results of the dynamic

OBA method with partial integration procedure along with the results of the SINS/GPS integration are presented in Fig. 27. The gyroscopes biases estimate results of the dynamic OBA method with full integration procedure along with the results of the SINS/GPS integration are presented in Fig. 28. We can draw the same conclusions from these results with those from the last experimental results. It should be noted that the gyroscopes biases in the two experiments are not the same with each other. This is because that the gyroscopes biases are always not the same after each application of power, which is known as the bias instability. Admittedly, this observation suggests that the gyroscopes biases should be estimated in each application, which can further give prominence to the significance of the proposed dynamic OBA method for the attitude alignment of the low-grade SINS.

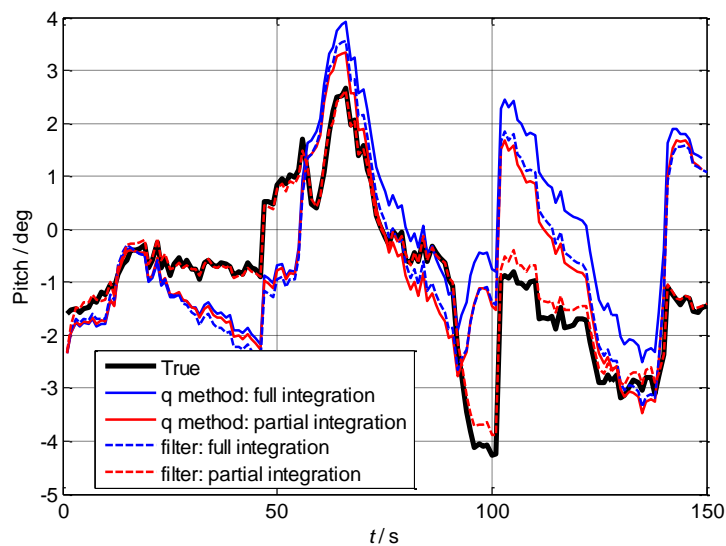

Fig. 21. Pitch angles estimates (second circle)

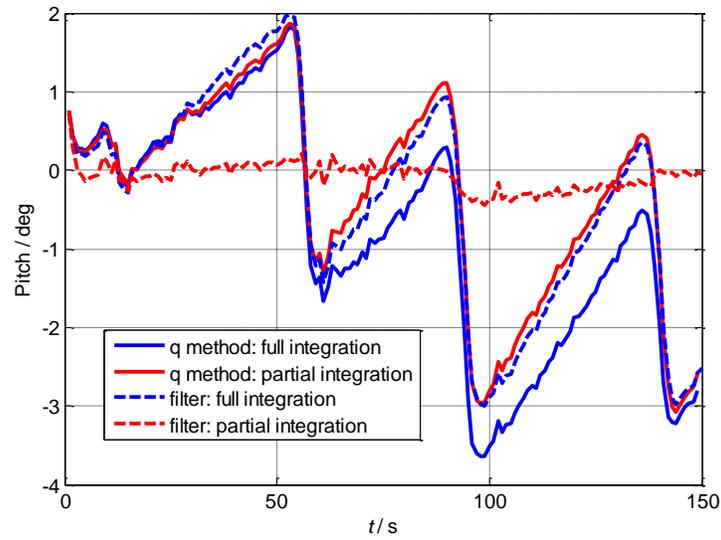

Fig. 22. Pitch angles estimate error (second circle)

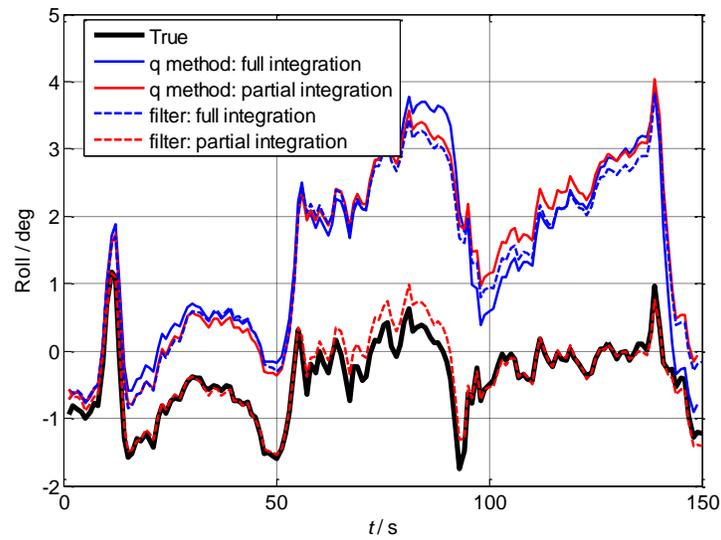

Fig. 23. Roll angles estimates (second circle)

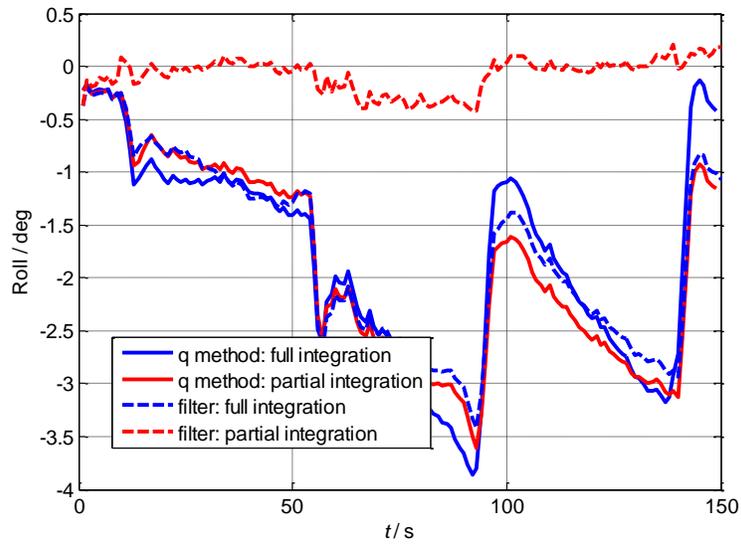

Fig. 24. Roll angles estimate error (second circle)

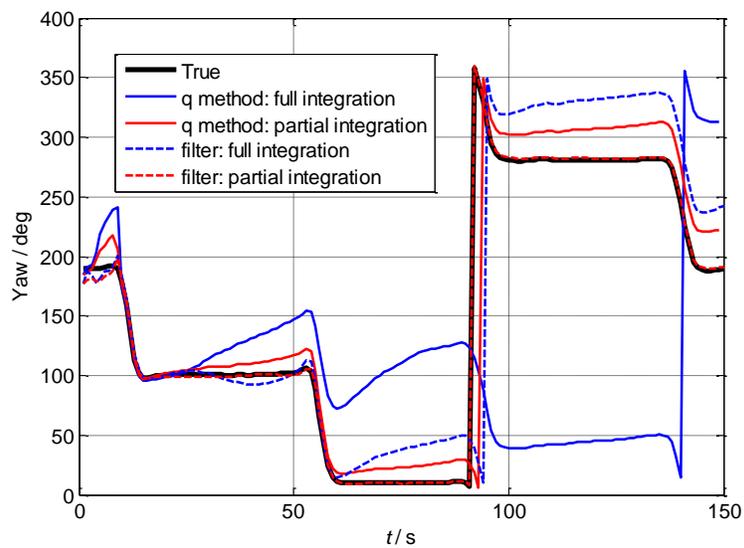

Fig. 25. Yaw angles estimates (second circle)

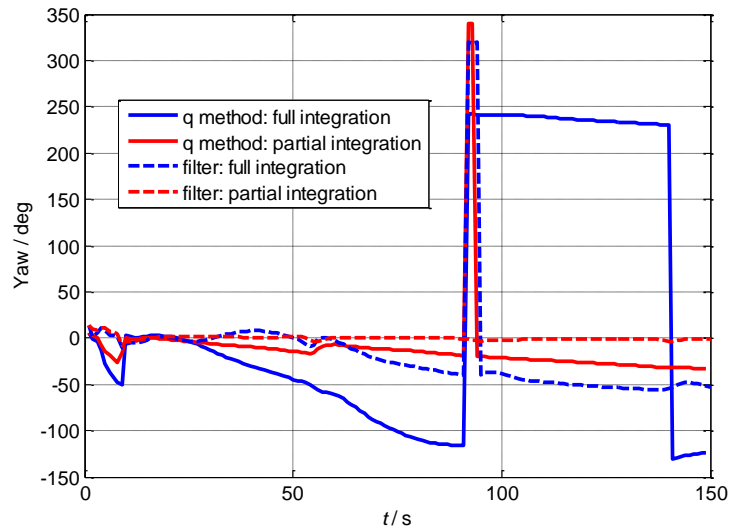

Fig. 26. Yaw angles estimate error (second circle)

Table V. ATTITUDE ERROR SUMMARY USING LOW-GRADE INERTIAL SENSORS (second circle, unit: deg)

|  | Pitch | Roll | Yaw |
| --- | --- | --- | --- |
| q method: full integration | -2.81 | -0.38 | -124.41 |
| q method: partial integration | -2.52 | -1.13 | -32.75 |
| filter: full integration | -2.47 | -1.01 | -51.71 |
| filter: partial integration | 0.07 | 0.18 | -1.47 |

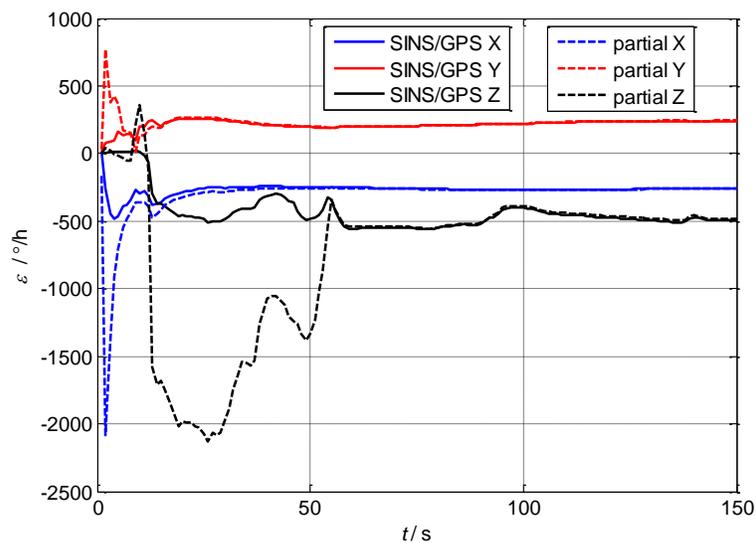

Fig. 27. Gyroscopes biases estimates of the "*filter: partial integration*" and SINS/GPS (second circle)

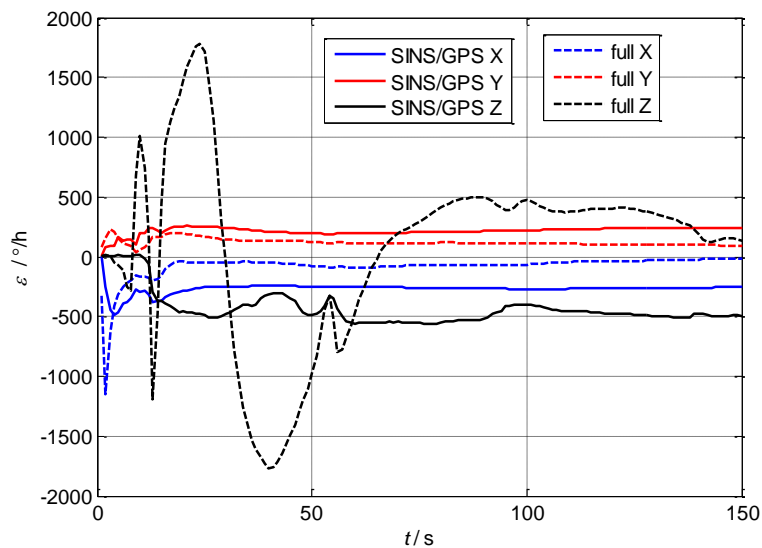

Fig. 28. Gyroscopes biases estimates of the "*filter: full integration*" and SINS/GPS (second circle)

It should be noted that the 100 interleaved samples used in the partial integration is only a special case. The length of integration time interval has a remarkable effect on the performance of the developed dynamic OBA method. For the navigation-grade SINS, making use of the sampling data more adequately can reach a better performance in terms of convergent speed and steady-state precision. For the low-grade SINS, however, the situation is a little complex. On the one hand, too less sampling data incorporated into the integration can not guarantee the appropriate convergent speed. On the other hand, too more sampling data incorporated into the integration may cause accumulative error due to the inherent inertial sensors' error. In this respect, it is desired to determine the length of integration time interval under different conditions. However, there is no universal method to achieve such end and it can only be determined through experiment study. Regarding the utilizing procedure of the sampling data over certain time intervals, the investigated method has a reasonable similarity to the finite impulse response (FIR) filter. Recently, Shmaliy and Simon have investigated unbiased FIR filter for the state-space model based problems and presented a unified forms for Kalman and FIR filtering [24, 25]. In [24], it is pointed out that the optimal window size can be easily estimated experimentally by minimizing an elaborate cost function without any "true" reference. The similarity between the Kalman and FIR filtering motives us to determine the length of integration time interval according to the proposed method in [24], which is the further research regarding the investigated dynamic OBA method.

## V. CONCLUSION

The newly developed OBA method has been proven to be an ingenious method for the SINS initial alignment under swaying or maneuvering conditions. Typically, there are mainly two integration procedures for the construction of the vector observations used in the OBA methods. However, the conflicting properties of the two integration procedures necessitate a further investigation of the OBA methods. Meanwhile, the existing OBA methods are all unable to estimate anything other than the attitude and can be viewed as static methods, which stultifies them for the low-grade SINS. Based on the construction process of the static OBA methods, a dynamic OBA method is developed to estimate the gyroscopes biases coupled with the attitude, which is expected to be applicable for the low-grade SINS. The existing static OBA methods and the developed dynamic OBA method with different integration procedures are comprehensively evaluated in terms of convergent speed and steady-state estimate using field test data from both navigation-grade and low-grade SINS. The comparison analysis shows that for the navigation-grade SINS the static OBA method is most celebrated due to its less computational cost and appropriate steady-state estimate precision. In this case, making use of more acceleration information can result in a much faster convergent speed with no degradation in steady-state estimate due to the high-grade specifications of the navigation-grade inertial sensors. For the low-grade SINS, the static OBA methods are no longer applicable due to the accumulative errors in the constructed vector observations caused by the large inertial sensors biases. In this case, the developed dynamic OBA method with interleaved integration procedure is most celebrated since it can estimate the gyroscopes biases appropriately and thus result in a much better attitude alignment performance. Making use of more acceleration information from the low-grade inertial sensors will cause much accumulative errors in the constructed vector observations, which can even stultify the dynamic OBA method.

A disadvantage of the proposed algorithm is that it can not estimate the accelerometer biases. Quite recently,

Wu et al. [23] propose an online constrained-optimization method to simultaneously estimate the attitude and the inertial sensor biases. This online constrained optimization method performs quite well in estimating the attitude and the accelerometer biases. However, its performance of estimating the gyroscopes biases is compromised due to the severe noise and disturbance. Therefore, this online constrained optimization method is confined to those accurate applications with navigation-grade SINS measurements [23]. In this respect, the online constrained optimization method and the proposed dynamic OBA method is complementary to each other in joint estimation of the gyroscopes/ accelerometer biases. In the future work, we will try to handle the gyroscopes/ accelerometer biases simultaneously by combining the virtue of the online constrained-optimization method and the proposed dynamic OBA method.

ACKNOWLEDGMENT

The authors would like to thank Gongmin Yan, and Zongwei Wu for providing the field test data.

smoothing," *Automatica*, vol. 49, no. 10, pp. 892–1899, 2013.

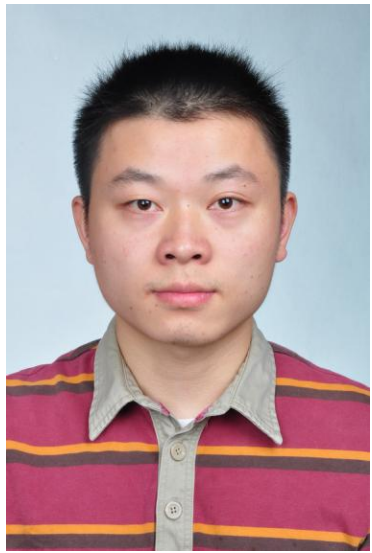

**Lubin Chang** (M'13) received B.Sc. and Ph.D. degrees in navigation from the Department of Navigation, Naval University of Engineering, Wuhan, P.R. China in 2009 and 2014, respectively.

He is currently a Lecturer with the Naval University of Engineering. His current research interests include inertial navigation systems, inertial-based integrated navigation systems, and state estimation theory.

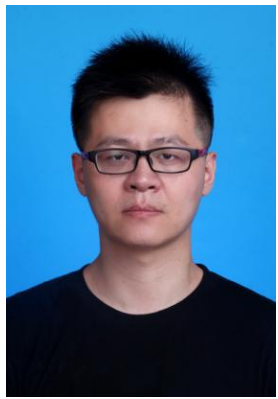

**Jingshu Li** received the B.Sc. and M.Sc. degrees in navigation from the Department of Control Engineering, Aviation University of Air Force, Changchun, P.R. China, in 2008 and 2010, respectively, and the Ph.D. degree in navigation from the Department of Navigation, Naval University of Engineering, Wuhan, P.R. China in 2014.

He is currently a Lecturer with the Naval University of Engineering. His current research interests include inertial navigation systems, and control algorithms.

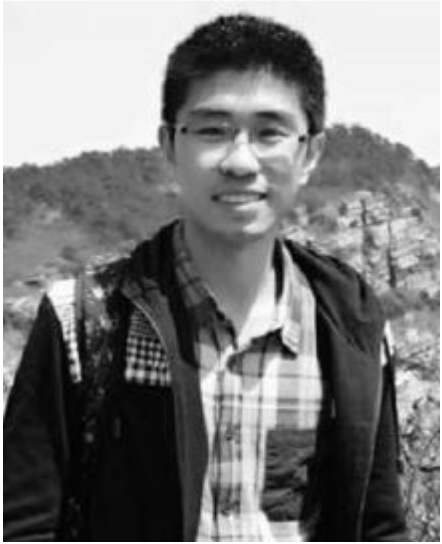

**Kailong Li** received B.Sc. and Ph.D. degrees in navigation from the Department of Navigation, Naval University of Engineering, Wuhan, P.R. China in 2011 and 2015, respectively.

He is currently a Lecturer with the Naval University of Engineering. His current research interests include inertial navigation systems, attitude estimation, and integrated navigation.